\pdfoutput=1

\documentclass[11pt]{article}

\usepackage{naacl2021}

\usepackage{times}
\usepackage{latexsym}

\usepackage[T1]{fontenc}

\usepackage[utf8]{inputenc}

\usepackage{microtype}

\usepackage{nicefrac} 
\usepackage{latexsym}
\usepackage{multirow}
\usepackage{booktabs}
\usepackage{amsmath}
\usepackage{booktabs}
\usepackage{subfig}
\usepackage{graphicx}
\usepackage{xspace}
\usepackage{times}
\usepackage{latexsym}
\usepackage{multirow}
\usepackage{algorithm,algorithmicx,algpseudocode}
\usepackage{url}
\usepackage{xcolor}
\usepackage[font={footnotesize}]{caption}
\usepackage{times}  
\usepackage{helvet} 
\usepackage{courier}  
\usepackage{booktabs}       
\usepackage{amsfonts}       

\DeclareMathOperator*{\argmax}{argmax}
\DeclareUnicodeCharacter{0307}{}

\definecolor{mygreen}{rgb}{0.64, 0.76, 0.68}
\definecolor{myyellow}{rgb}{0.98, 0.94, 0.75}
\definecolor{mygreen}{rgb}{0.68, 0.9, 0.6}

\definecolor{myblue}{rgb}{0.9, 0.1, 0.94}
\newcommand{\modelname}{\textit{Refactor}\xspace}
\newcommand{\singlestep}{Supervised\xspace}

\title{\textit{RefSum}: Refactoring Neural Summarization}

\author{Yixin Liu, Zi-Yi Dou, Pengfei Liu \thanks{\ \  Corresponding author.}\\
  Carnegie Mellon University \\
  \texttt{\{yixinl2,zdou,pliu3\}@cs.cmu.edu}}

\begin{document}
\maketitle
\begin{abstract}
Although some recent works show potential complementarity among different state-of-the-art systems, few works try to investigate this problem in text summarization.
Researchers in other areas commonly refer to the techniques of \textit{reranking} or \textit{stacking} to approach this problem.
In this work, we highlight several limitations of previous methods, which motivates us to present a new framework \textit{Refactor} that provides a unified view of text summarization and summaries combination.
Experimentally, we perform a comprehensive evaluation that involves twenty-two base systems, four datasets, and three different application scenarios.
Besides new state-of-the-art results on CNN/DailyMail dataset (46.18 ROUGE-1), we also elaborate on how our proposed method addresses the limitations of the traditional methods and the effectiveness of the \modelname model sheds light on insight for performance improvement.
Our system can be directly used by other researchers as an off-the-shelf tool to achieve further performance improvements. We open-source all the code and provide a convenient interface to use it: \url{https://github.com/yixinL7/Refactoring-Summarization}.
We have also made the demo of this work available at: \url{http://explainaboard.nlpedia.ai/leaderboard/task-summ/index.php}. 
\end{abstract}

\section{Introduction}
\label{sec:intro}
In neural text summarization, system designers commonly have flexible choices in model architectures \cite{rush-etal-2015-neural,kedzie-etal-2018-content}, decoding strategies \cite{paulus2018deep} (e.g. beam search)
and etc.
As a result, even on the same dataset, different selection biases of these choices will lead to diverse system outputs~\cite{kedzie-etal-2018-content,hossain-etal-2020-simple}.

To combine complementarity of system's output under different setups, researchers have made some preliminary efforts on two-stage learning \cite{collins-koo-2005-discriminative, huang-2008-forest,gonzalez-rubio-etal-2011-minimum,mizumoto-matsumoto-2016-discriminative}, consisting of
(i) a \textit{base-stage}: first generates different outputs under different setups, and (ii) a \textit{meta-stage}: then aggregates them in diverse ways, exemplified by \textit{stacking} that uses a high-level model to combine \textit{multiple} low-level models \cite{DBLP:conf/ijcai/TingG97}, or \textit{reranking} \cite{collins-koo-2005-discriminative}, which aims to rerank different outputs of \textit{one} system.
Although these methods each play a role in different scenarios, they suffer from following potential limitations:

\begin{figure}
    \centering
    \includegraphics[width=0.98\linewidth]{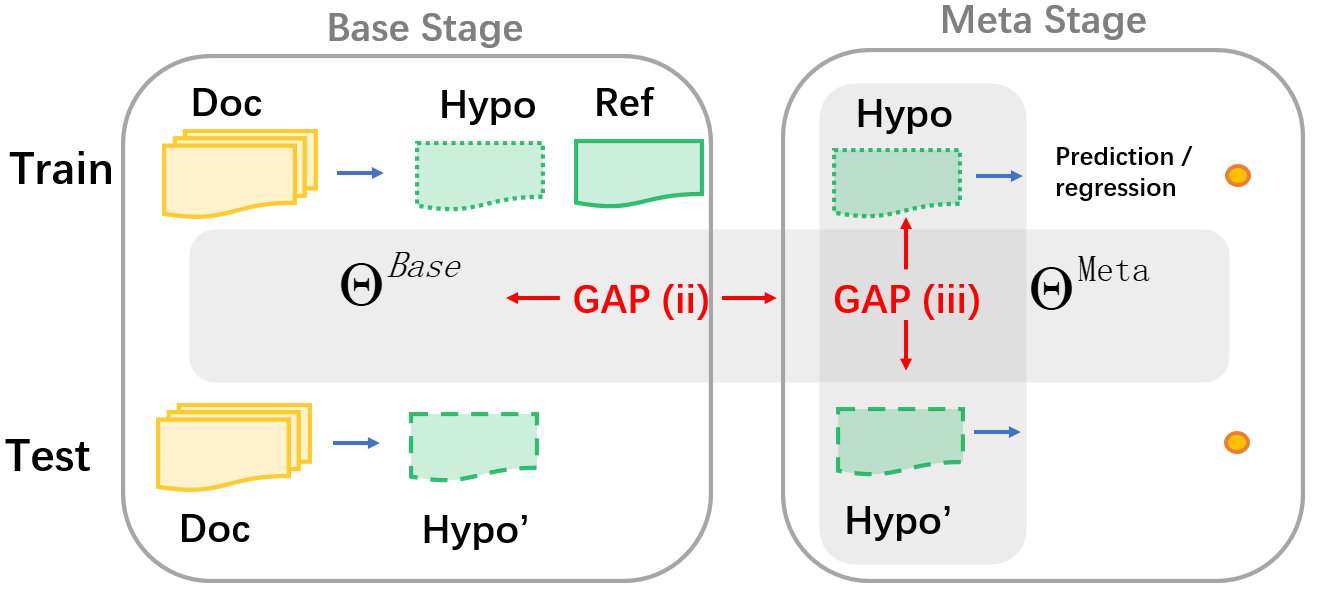}
    \caption{Illustration of two-stage learning. ``\texttt{Doc}, \texttt{Hypo}, \texttt{Ref}'' represent ``\texttt{input document}, \texttt{generated hypothesis}, \texttt{gold reference}'' respectively. ``\texttt{Hypo'}'' represents texts generated during test phase. $\Theta^{\text{Base}}$ and $\Theta^{\text{Meta}}$ represent learnable parameters in two stages.}
    \label{fig:intro}
\end{figure}

(i) \textit{Ad-hoc Methods}: most existing methods are designed for a specific scenario.
For example, \citet{li-etal-2015-improving-update} and \citet{narayan-etal-2018-ranking} resort to reranking techniques to select summary-worthy sentences that are usually generated from one system.
By contrast, \citet{hong-etal-2015-system} focus on summaries generated from different systems and use a non-neural system combination method to make their complementary advantages.
Few works explore if the complementarity existing in different scenarios could be utilized in a unified framework.

(ii) \textit{Base-Meta Learning Gap}: parameterized models between two learning stages are relatively independent. 
For example, \citet{zhou-etal-2017-neural} and \citet{ DBLP:conf/ijcai/HuangZTWLXSL20} adapt the seq2seq \cite{sutskever2014sequence} framework as the meta model for combination, which takes the outputs of multiple base systems as a part of the inputs for machine translation. 
As a result, there is no parameter sharing between the meta model and base systems as shown in Fig.~\ref{fig:intro}, which prevents the meta model from fully utilizing the knowledge encoded in the base systems.

(iii) \textit{Train-Test Distribution Gap}: regarding the meta-learning stage, there is a distribution gap between the training and test distributions. 
Fig.~\ref{fig:intro} elucidates this phenomenon:
the training distribution of \texttt{Hypo} differs from the test distribution of \texttt{Hypo'}. Although both two are outputs from the base stage, \texttt{Hypo} would be more accurate (closer to gold summaries) since it is the output during the training phase.

In this work, we aim to address these limitations by proposing a general framework, named \modelname
, which can not only serve as a base system to construct a summary by selecting sentences from the source document but also act as a meta system to select the best system output from multiple candidates.
The unification of base and meta systems allows them to share a set of parameters, thereby alleviating the ``Base-Meta learning gap''.
Besides, we propose a \textit{pretrain-then-finetune} paradigm for \modelname that mitigates the ``Train-Test distribution gap''.
In practice, our proposed \modelname can be applied to different scenarios. 
For example, 
as a meta system, it can be used for multiple system combination or single system re-ranking. 

Our contributions can be briefly summarized as:

(1) We dissect two major factors that influence the performance of two-stage learning when leveraging the complementarity among different systems: (i) Base-Meta Learning Gap (ii) Train-Test Distribution Gap;

(2) We show these two types of gaps can be alleviated by promoting communication between the two stages in \S\ref{sec:refactoring} 
, and therefore present a new paradigm where the base and meta learners are parameterized with shared parameters;

(3) We have made comprehensive experiments (twenty-two top-scoring systems, four datasets).
In addition to achieving state-of-the-art results on CNN/DailyMail dataset (\S\ref{sec:exp}) by a significant margin, the efficacy of the proposed \modelname opens up a thought-provoking direction for performance improvement: instead of pursuing a purely end-to-end system, a promising exploration is to incorporate different types of inductive biases stage-wisely with the same parameterized function.
Our experimental results demonstrate that there exists complementarity introduced by decoding algorithms (e.g. beam search) \S\ref{subsec:cnndm-single} or system combination \S\ref{subsec: stacking} among the current state-of-the-art summarization systems, which can be effectively utilized by our model for boosting the system performance.

 \section{Preliminaries}
Existing works commonly design systems in an end-to-end fashion \cite{sutskever2014sequence,sukhbaatar2015end}, which, though effective, also proves to be insufficient in some scenarios \cite{glasmachers2017limits,webb2019ensemble}. 
Instead of optimizing a system in an end-to-end fashion, one more flexible paradigm, stage-wise learning, is to break down the holistic process into different stages. 
The basic idea is to incorporate different types of inductive biases stage-wisely and two typical examples are: \textit{Stacking} and \textit{Reranking}.

\paragraph{Stacking}

Stacking (a.k.a, Stacked Generalization) is a general method of
using a high-level model to combine lower-level models to achieve greater predictive accuracy \cite{DBLP:conf/ijcai/TingG97}.
In NLP research, this method has been widely explored in machine translation (MT) task. Traditionally, it is used to improve the performance of statistical MT systems~\citep{gonzalez-rubio-etal-2011-minimum, watanabe-sumita-2011-machine, duh2011generalized, mizumoto-matsumoto-2016-discriminative}. Some recent work~\citep{zhou-etal-2017-neural, DBLP:conf/ijcai/HuangZTWLXSL20} also extends this method to neural MT where the meta model and base systems are all neural models.
There is a handful of works about system combination for summarization \cite{hong-etal-2015-system}, in which a feature-based meta model is used for combining unsupervised text summarization systems.

\paragraph{Reranking}

Reranking
is a technique to improve performance by reranking the output of an existing system, which has been
widely used across different NLP tasks, such as constituency parsing~\citep{collins-koo-2005-discriminative, huang-2008-forest}, dependency parsing~\cite{zhou-etal-2016-search, do-rehbein-2020-neural}, semantic parsing~\citep{ge-mooney-2006-discriminative, yin-neubig-2019-reranking}, machine translation~\citep{shen2004discriminative, mizumoto-matsumoto-2016-discriminative}.

Comparing \textit{reranking} and \textit{stacking}, both of them involve two-stage learning and the first stage would provide multiple candidate outputs as the input for the second stage.
However, they differ in the way how multiple candidate outputs are generated at the first stage. Specifically, reranking usually decodes $k$-most qualified results during inference, using one base system. By contrast, stacking generates multiple outputs that are usually from different base systems.

\section{Summarization as Two-stage Learning}
In what follows, we detail how to formulate summarization as a two-stage learning task.

\paragraph{Base system}
The system in the base stage aims to generate a summary based on the input text.
Specifically, given a document $D = \{s_1, \cdots, s_n\}$ with $n$ sentences,
we refer to $C$ as a \textit{candidate summary} of $D$ generated by a summarization system, which can be parameterized in diverse forms:
\begin{align}
    C = \textsc{Base}(D, \mathcal{T}, \mathcal{S},\Theta^{\text{base}})
 \end{align}
where $\textsc{Base}(,\Theta^{\text{base}})$ represents a base system that can be instantiated either as an extractive model or abstractive model with a specific experimental setup: training method $\mathcal{T}$, decoding strategy $\mathcal{S}$.

\paragraph{Meta system}
In practice, different choices of parameterized function $\textsc{Base}(\cdot)$, training method $\mathcal{T}$ and decoding strategy $\mathcal{S}$ commonly lead to different candidate summaries, $\mathcal{C} = \{C_1, \cdots, C_k\}$, where $\mathcal{C}$ represents a set of different candidate summaries.
The goal of the meta system is to utilize complementarities among $\mathcal{C}$ by popular techniques, such as reranking and system combination.

Specifically, given a set of candidate summaries $\mathcal{C}$, a meta system is used to re-construct a new candidate summary $C^{*}$
\begin{align}
    C^{*} = \textsc{Meta}(D, \mathcal{C}, \Theta^{\text{meta}})
\end{align}
where $\Theta^{\text{meta}}$ represents learnable parameters of the meta system.

\section{Refactoring Text Summarization}
\label{sec:refactoring}
Despite effectiveness of existing meta systems, they, as briefly mentioned in \S\ref{sec:intro}, suffer from two major problems: (i) \textit{Base-Meta Learning Gap} and 
(ii) \textit{Train-Test Distribution Gap}.

\subsection{Refactoring}

In this paper,  we propose the model \textbf{\modelname} that unifies the goal of the base and meta systems by the view that a summary can be generated by selecting the best combination of document sentences.
Therefore, both base and meta systems aim to select an optimal candidate summary, and they only differ in how the candidate summary set is constructed. 
For example, \modelname can be a base system when the candidate summary set $\mathcal{C}$ is formed by directly enumerating different combinations of \textit{document sentences} and would be a meta system when $\mathcal{C}$ represents summaries from different systems.
This formulation is advantageous in two points:

(1) No matter where a system selects (from document sentences or multiple system outputs), the chosen criteria that define a good summary are shared. 
Therefore, the learning process of base and meta systems can be parameterized using a set of parameters, maximizing the information-sharing across two stages and mitigating the \textit{Base-Meta Learning Gap}.
\begin{align}
    C^{*} = \textsc{Refactor}(D, \mathcal{C}, \Theta^{\text{refactor}}),
\end{align}
where $\textsc{Refactor}(\cdot, \Theta^{\text{refactor}})$ is the \modelname model, and the candidate summaries $\mathcal{C}$ can be constructed in different ways.

(2) Additionally, learning to select candidate summaries from document sentences enables the system to see more diverse candidates with different distributions.
This is effective for solving the \textit{Train-Test Distribution Gap}, where the distribution of the meta system outputs in training samples deviates from the test one.

Specifically, our proposed \modelname first learns to select candidate summaries from document sentences (pre-trained \modelname) and then learns to select candidate summaries from different system outputs (fine-tuned \modelname). 

\subsection{Pre-trained Refactor}

Pre-trained \modelname takes as input a document $D = \{s_1, \cdots, s_n\}$ as well as a set of candidate summaries $\mathcal{C} = \{C_1, \cdots, C_m\}$, which can be constructed by enumerating possible combinations of source sentences with heuristic pruning.
For example, an extractive system could be used to prune unlikely sentences to control the number of candidates.
$\textsc{Refactor}(\cdot, \Theta^{\text{refactor}})$ is instantiated as a score function which quantifies the degree to which a candidate summary $C_i$ is matched with the source document $D$.
\begin{align}
    C^{*} &= \textsc{Refactor}(D, \mathcal{C}, \Theta^{\text{refactor}}) \nonumber \\
    &= \argmax_{C_i \in \mathcal{C}}(\textsc{Score}(\mathbf{D}, \mathbf{C}_i))
\end{align}
where $\mathbf{D}$ and $\mathbf{C}_i$ denote document and summary representations respectively, which are calculated by a BERT~\citep{devlin2019bert} model. $\textsc{Score}(\cdot)$ is a function that measures the similarity between a document and candidate summary.

\paragraph{Contextualized Similarity Function}

To instantiate $\textsc{Score}(\cdot)$, we follow the forms as mentioned in \citet{zhang2019bertscore,zhao-etal-2019-moverscore,gao-etal-2020-supert}, which have shown superior performance on measuring semantic similarity between documents and summaries.

Specifically, $\textsc{Score}(\cdot)$ is defined based on the greedy matching algorithm, which matches every word in one text sequence to the most similar word in another text sequence and vise versa. Given the document embedding matrix $\mathbf{D} = 
\langle \mathbf{d}_1, \cdots, \mathbf{d}_k \rangle$ and the candidate embedding matrix $\mathbf{C} = 
\langle \mathbf{c}_1, \cdots, \mathbf{c}_l \rangle$ encoded by BERT, 
$\textsc{Score}(\cdot)$ can be calculated as:
\begin{equation}
     \textsc{Score}(\mathbf{D}, \mathbf{C}) = 2 \frac{\mathrm{R}(\mathbf{D}, \mathbf{C}) \cdot \mathrm{P}(\mathbf{D}, \mathbf{C})}{\mathrm{R}(\mathbf{D}, \mathbf{C}) + \mathrm{P}(\mathbf{D}, \mathbf{C})}
\end{equation}
where the weighted recall $\mathrm{R}$, precision $\mathrm{P}$ are defined as follows:\footnote{We found that adding $1$ to the precision and recall helps to stabilize the training.}
\begin{align}
        \mathrm{R}(\mathbf{D}, \mathbf{C}) &= \frac{\sum_i w_i \max_j \mathrm{cos}(\mathbf{d}_i, \mathbf{c}_j)}{\sum_{i} w_i} + 1, \\
    \mathrm{P}(\mathbf{D}, \mathbf{C}) &= \frac{\sum_j \max_i \mathrm{cos}(\mathbf{d}_i, \mathbf{c}_j)}{l} + 1,
\end{align}

$w_i$ is the weight of the $i$-th token in the document. 
We use weighted recall $\mathrm{R}$ based on the assumption that for text summarization, tokens in the source document have different importance and the summary should capture the most important information of the source document. 
Therefore, we introduce a weighting module built by a two-layer Transformer~\citep{NIPS2017_7181} assigning weights $w_i$:
\begin{equation}
    w_i = \frac{\exp{({\mathrm{dot}({\textbf{d}}_i, \hat{\textbf{d}}_0)}/{\sqrt{d}})}}{\sum_j\exp({\mathrm{dot}({\textbf{d}}_j, \hat{\textbf{d}}_0)}/{\sqrt{d}})},\vspace{-3pt}
\end{equation}
where $\hat{\mathbf{D}} = \mathrm{Transformer}(\mathbf{D})$
and  $\hat{\textbf{d}}_0 = \hat{\mathbf{D}}[0]$ represents the embedding of the ``\texttt{[CLS]}'' token which encodes the global information. $d$ is the dimension of $\textbf{d}_i$.

\paragraph{Learning Objective}
We use a ranking loss to learn the parameter $\Theta^{\text{refactor}}$, inspired by the assumption \citep{zhong-etal-2020-extractive} that a good candidate summary should be as close with the source document as possible.
Formally, 
\begin{equation}
\label{eq:loss_c}
\begin{split}
      L = &\sum_i \sum_{j > i} \max(0, \textsc{score}(\mathbf{D}, \mathbf{C}_j) \\
      & - \textsc{score}(\mathbf{D}, \mathbf{C}_i)  + (j - i) * \lambda_{c})\\
\end{split}
\end{equation}
where $C_i$ and $C_j$ denote the $i$-th and $j$-th sample of the candidate list which is descendingly sorted by the ROUGE~\citep{lin-2004-rouge} scores between the reference summary $\hat{C}$ and candidates. That is, $\mathrm{ROUGE}(C_i, \hat{C}) > \mathrm{ROUGE}(C_j, \hat{C})$ for $i < j$.
$\lambda_c$ is the corresponding margin set to $0.01$.

\subsection{Fine-tuned Refactor}
In order to fit the distributions of the specific types of input, we then fine-tune \modelname using the outputs generated by the base systems.
Specifically, fine-tuning is also based on Eq.~\ref{eq:loss_c} where the candidate summaries $C$ are generated by the base systems under different application scenarios. 
\begin{figure}[t!]
    \centering
    \includegraphics[width=0.8\linewidth]{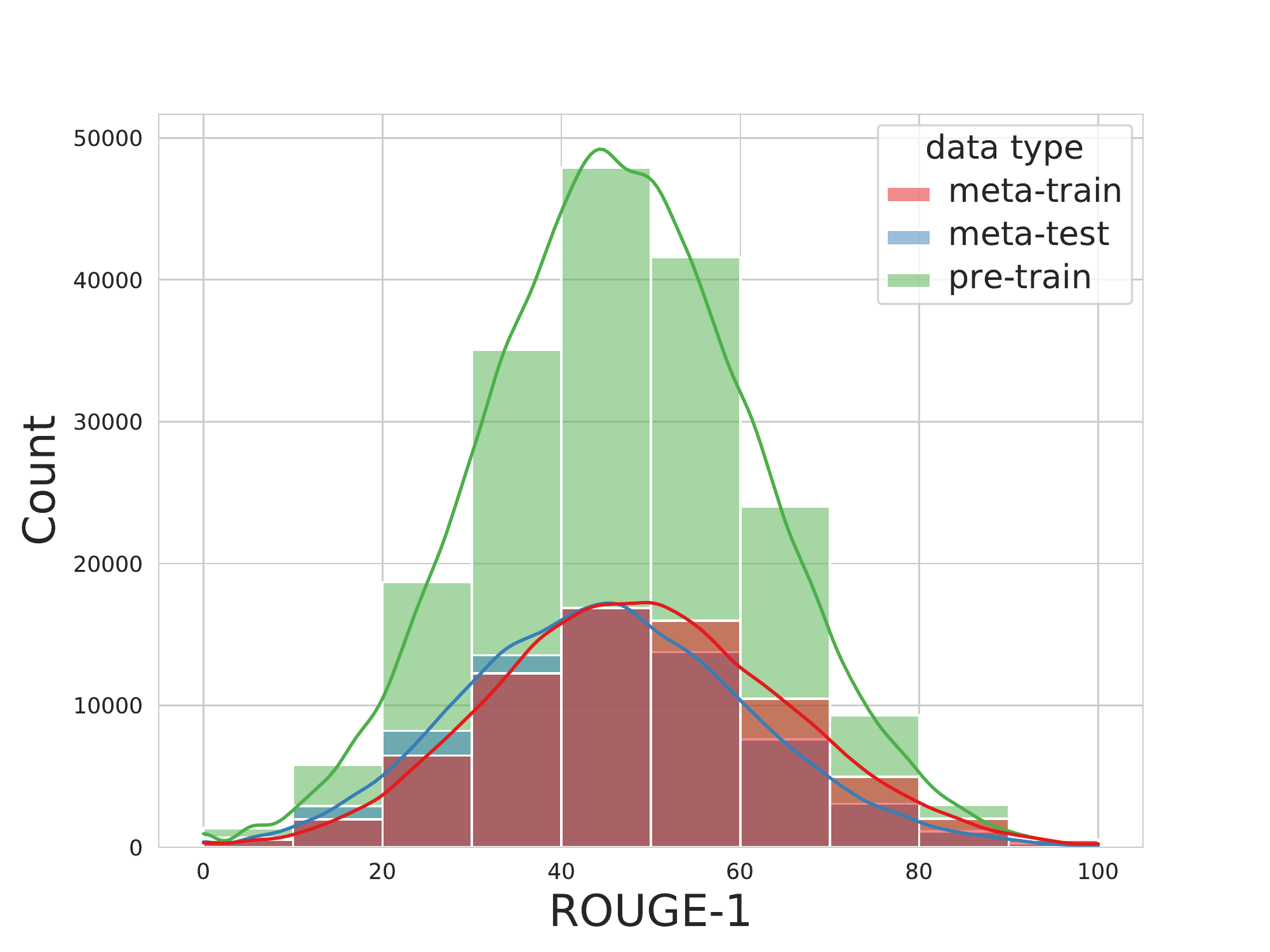}
    \caption{ROUGE-1 distributions of the candidates in pre-training stage training set (\textit{pre-train}), fine-tuning stage training set (\textit{meta-train}) and fine-tuning stage test set (\textit{meta-test}) on \texttt{XSum} dataset.} 
    \label{fig:dist-xsum}
\end{figure}
\paragraph{Why does \textit{Pre-train and Fine-tune} matter?}

We elaborate on the proposed two-step training using a real case.
Fig.~\ref{fig:dist-xsum} depicts the distribution of ROUGE-1 scores regarding the candidate summaries in the pre-training stage training set, fine-tuning stage training set   and test set   on the \texttt{XSum} dataset, where we sample the same number of $\{\textrm{document}, \textrm{candidate summaries}\}$ pairs.
We can observe that:

(i) there is a distribution gap between train and test samples in fine-tuning stage.
(ii) in \textit{pre-training} stage the pre-trained \modelname~has seen a large number of candidate summaries with diverse performance (ROUGE value), which improves its generalization ability.
In \S\ref{sec:exp} we will show that the Pre-train and Fine-tune paradigm outperforms one-step training where the model is directly trained with data generated from the base systems.

\subsection{Application Scenarios}
\label{sec:scenario}

Our \modelname can be used as different roles in different scenarios as follows.

\subsubsection{Refactor as Base Learner}
The pre-trained \modelname can not only be fine-tuned for a better selection of candidate summaries, but also be regarded as a base system, providing one system output.
This feature of \modelname maximizes parameter sharing across the two training stages.

\subsubsection{Refactor as Meta Learner}
Both pre-trained \modelname and fine-tuned \modelname can be used as a meta system to select the best candidate when we have multiple system summaries. 
In this work, we explore the following settings:

\label{subsed: single}
(1) \textbf{Single System}:
It considers re-ranking candidate summaries generated from a single abstractive system using beam search. 

(2) \textbf{Multi-system Summary-level}:
It is tasked to select the best candidate summary from the results of different systems.

(3) \textbf{Multi-system Sentence-level}:
We also take a step towards the fine-grained fusion of summaries from extractive and abstractive systems.
Specifically, here candidate summaries are generated by combining the results of different systems at the sentence level.

\section{Experiments}
\label{sec:exp}
\subsection{Datasets}

We mainly experiment on four datasets, whose statistics are shown in Tab.~\ref{tab:data}.

\noindent \textbf{CNNDM}\footnote{\url{https://cs.nyu.edu/~kcho/DMQA/}}~\citep{hermann2015teaching} is a widely used dataset containing news articles and the associated highlights which are used as the reference summaries. 
We follow the work of~\citet{nallapati-etal-2016-abstractive} for data preprocessing.

\noindent \textbf{XSum}\footnote{\url{https://github.com/EdinburghNLP/XSum}}~\citep{narayan2018don} contains online articles collected from BBC with highly abstractive one-sentence summaries.

\noindent \textbf{PubMed}\footnote{\url{https://github.com/acohan/long-summarization}} ~\citep{cohan-etal-2018-discourse} contains scientific papers collected from PubMed.com.

\noindent \textbf{WikiHow}\footnote{\url{https://github.com/mahnazkoupaee/WikiHow-Dataset}}~\citep{DBLP:journals/corr/abs-1810-09305} is a large-scale dataset constructed from the articles using online WikiHow knowledge base.

\begin{table}[t]
  \centering
  \small
    \begin{tabular}{@{\extracolsep{1pt}}lccccc}
    \toprule
    \multirow{2}{*}{Datasets} & \multicolumn{3}{c}{\# Num} & \multicolumn{2}{c}{Avg. Len} \\
    \cmidrule{2-4} \cmidrule{5-6}
    & Train & Valid & Test & Doc. & Sum. \\
    \midrule
    CNNDM & 287K & 13K & 11K & 768.6 & 55.7 \\
    XSum & 203K & 11K & 11K & 429.2 & 23.3 \\
    PubMed & 83K & 4.6K & 5K & 468.7 & 210.3 \\
    WikiHow & 168K & 6K & 6K & 579.1 & 62.2 \\
    \bottomrule
    \end{tabular}%
  \caption{Datasets Statistics. Len is the length of tokens.}
  \label{tab:data}%
\end{table}%

\subsection{Base Systems}
Below, we mainly use BART, GSum and PEGASUS as the base systems since they have achieved state-of-the-art performance on at least one dataset.

\noindent\textbf{BART}~\citep{lewis-etal-2020-bart} is a large pre-trained sequence-to-sequence model that achieves strong performance on the abstractive summarization. 

\noindent\textbf{GSum}~\citep{dou2020gsum} enhances the performance of BART using additional guidance information, which achieves the current state-of-the-art performance on the \texttt{CNNDM} dataset.

\noindent\textbf{PEGASUS}~\citep{pmlr-v119-zhang20ae} 
achieves competitive performance on various summarization datasets and is the current state-of-the-art on the \texttt{XSum} dataset.

To make a comprehensive evaluation of our proposed model, we additionally collect 19 top-scoring systems as base systems on \texttt{CNNDM}.\footnote{Since \texttt{CNNDM} is the most popular dataset, we can collect more existing systems on it.}
In details, for \S\ref{subsec:exp-19} we use the following systems: pointer-generator+coverage~\citep{see-etal-2017-get}, REFRESH~\citep{narayan-etal-2018-ranking}, fastAbsRL-rank~\citep{chen-bansal-2018-fast}, CNN-LSTM-BiClassifier~\citep{kedzie-etal-2018-content}, CNN-Transformer-BiClassifier~\citep{zhong-etal-2019-searching}, CNN-Transformer-Pointer~\citep{zhong-etal-2019-searching},
BERT-Transformer-Pointer~\citep{zhong-etal-2019-searching},
Bottom-Up~\citep{gehrmann-etal-2018-bottom}, NeuSum~\citep{zhou-etal-2018-neural}, BanditSum~\citep{dong-etal-2018-banditsum}, twoStageRL~\citep{zhang-etal-2019-pretraining},
preSummAbs~\citep{liu-lapata-2019-text}, preSummAbs-ext~\citep{liu-lapata-2019-text},
HeterGraph~\citep{wang-etal-2020-heterogeneous},
MatchSum~\citep{zhong-etal-2020-extractive},
Unilm-v1~\citep{dong2019unified}, Unilm-v2~\citep{dong2019unified},
T5~\citep{JMLR:v21:20-074}.

\subsection{Baseline Systems}

\noindent \textbf{Neural system combinator}:
We use BERTScore~\citep{zhang2019bertscore} as an unsupervised baseline with neural models, which is an automatic evaluation metric computing the similarity of text pairs based on the corresponding BERT-encoded representations. We use it to directly compute the similarity score between the source documents and candidate summaries.

\noindent \textbf{Non-Neural system combinator}:
We use RankSVM\footnote{\url{http://www.cs.cornell.edu/people/tj/svm\_light/svm\_rank.html}}~\citep{joachims2002optimizing} as a non-neural baseline. 
We perform cross-validation on the development set for hyper-parameter searching and train the model on the development set. The set of features is listed in Appendix~\ref{sec:ranksvm}.

\noindent \textbf{Oracles}:
We compare our model with sample-wise \textbf{Min}, \textbf{Max} and \textbf{Random} oracles using ROUGE.

\subsection{Training Details}
For the following experiments in \S\ref{subsec:cnndm-single}, \S\ref{subsec: stacking} and \S\ref{subsec:exp-19} on \texttt{CNNDM}, we pre-train the \modelname model with a candidate set generated by enumerating combinations of sentences in the source documents. 
To reduce the number of candidates, we prune the sentences assigned with lower scores by an extractive model, BERTSum~\citep{liu-lapata-2019-text}, following~\citet{zhong-etal-2020-extractive}.
The maximum number of candidates for one data sample is 20.
The \textbf{\textit{pre-trained} \modelname} is also used a base system in \S\ref{subsec: stacking}, whose outputs are used together with other base systems as candidate summaries.
For different experiments, we fine-tune \textit{pre-trained} \modelname on the base system's output, and name the model as \textbf{\textit{fine-tuned} \modelname}.
To analyze the effectiveness of the proposed two-stage training, we additionally train the model without the pre-training step, which is named as \textbf{\textit{supervised} \modelname}.

The pre-trained BERT model we used is from \textit{Transformers} library~\citep{wolf-etal-2020-transformers}.\footnote{We use the `bert-base-uncased' version with 110M parameters.}
We use Adam optimizer~\citep{DBLP:journals/corr/KingmaB14} with learning rate scheduling.
\begin{align}
    lr &= 0.002 \cdot \min(\mathrm{step\_num}^{-0.5}, \\ \nonumber &\mathrm{step\_num}\cdot\mathrm{warmup\_steps}^{-1.5}),
\end{align}
where the $\mathrm{warmup\_steps}$ is 10000.
The model performance on the validation set is used to select the checkpoint.
Pre-training takes around 40 hours on 4 GTX-1080-Ti GPUs while fine-tuning takes around 20 hours.

\begin{table}[t!]
\small
\centering
\begin{tabular}{lllll}
\toprule
\textbf{System} & \textbf{Method} & \textbf{R-1} & \textbf{R-2} & \textbf{R-L}\\
\midrule 
 \multirow{9}{*}{BART} & Base & 44.26 & 21.12 & 41.16 \\
 &Min & 41.58 & 19.27 & 38.69 \\
 &Max & 47.22 & 23.28 & 43.90 \\
 &Random & 44.40 & 21.26 & 41.28 \\
 &BERTScore & 44.50 & 21.28 & 41.37 \\
 &RankSVM & 44.50 & 21.39 & 41.43 \\
\cmidrule{2-5}
 &Supervised\dag & 45.05 & 21.64 & 41.92 \\
 &Pre-trained\dag & 44.78 & 21.49 & 41.68\\ 
 &Fine-tuned\dag & \textbf{45.15} & \textbf{21.70} & \textbf{42.00} \\
\midrule 
 \multirow{9}{*}{GSum} & Base & 45.93 & 22.30 & 42.68 \\
 &Min & 44.37 & 21.25 & 41.29 \\
 &Max & 47.37 & 23.21 & 43.99 \\
 &Random & 45.84 & 22.22 & 42.61 \\
 &BERTScore & 45.84 & 22.25 & 42.64 \\
 &RankSVM & 46.04 & 22.29 & 42.78 \\
\cmidrule{2-5}
 & \singlestep\dag & 46.11 & 22.32 & 42.85 \\
 &Pre-trained & 45.88 & 22.23 & 42.67 \\ 
 &Fine-tuned\dag & \textbf{46.18} & \textbf{22.36} & \textbf{42.91} \\
\bottomrule
\end{tabular}
\caption{\label{tab:single} Single system reranking on \texttt{CNNDM}. \textbf{Base} denotes the base system. 
\textbf{Supervised} denotes the \modelname directly trained on the base systems' outputs. \textbf{Pre-trained} denotes the pre-trained \modelname. \textbf{Fine-tuned} denotes the fine-tuned model. 
R-1, R-2 and R-L denote ROUGE-1, ROUGE-2 and ROUGE-L. \dag: significantly better than the base system ($p < 0.01$).}
\end{table}
\subsection{Exp-I: Single System Reranking} 
\label{subsec:cnndm-single}

We use BART and GSum for this experiment, 
and use beam search to generate the candidate summaries where the beam size is set to 4.

The results are listed in Tab.~\ref{tab:single}, which shows that (1)  \modelname can boost the base system's performance by a significant margin, 
(2) the \textit{fine-tuned} \modelname outperforms \textit{supervised} \modelname directly trained on the base system's outputs, showing the effectiveness of the two-step training.
Notably, we observe the fine-tuned \modelname can boost BART's performance from 44.26 to 45.15 on ROUGE-1, indicating that the top-$1$ output selected by beam search is not always the best one, and \modelname can effectively utilize the complementarity introduced by considering all the beam search results.

\begin{table}[t!]
\small
\centering
\begin{tabular}{lllll}
\toprule
\textbf{Setting} & \textbf{Method} & \textbf{R-1} & \textbf{R-2} & \textbf{R-L}\\
\midrule
\multirow{3}{*}{Base} & BART & 44.26 & 21.12 & 41.16 \\
& Refactor & 44.13 & 20.51 & 40.29 \\
& GSum & 45.93 & 22.30 & 42.68 \\
\midrule
\multirow{8}{*}{Two} & Min & 40.40 & 17.64 & 37.12\\
& Max & 47.99 & 23.99 & 44.33 \\
& Random & 44.25 & 20.87 & 40.78 \\
& BERTScore & 43.95 & 20.45 & 40.23 \\
& RankSVM & 44.66 & 21.32 & 41.44 \\
\cmidrule{2-5}
& \singlestep\dag & 44.75 & 21.40 & 41.47 \\
& Pre-trained\dag & 44.66 & 21.19 & 41.15 \\ 
& Fine-tuned\dag & \textbf{45.04} & \textbf{21.61} & \textbf{41.72} \\
\midrule 
\multirow{8}{*}{Three} & Min & 39.51 & 17.01 & 36.35 \\
& Max & 49.94 & 25.59 & 46.30 \\
& Random & 44.82 & 21.35 & 41.44 \\
& BERTScore & 44.10 & 20.64 & 40.42 \\
& RankSVM & 45.72 & 22.13 & 42.58 \\
\cmidrule{2-5}
& \singlestep & 45.80 & 22.25 & 42.68 \\
& Pre-trained & 45.27 & 21.74 & 41.93 \\
& Fine-tuned\dag & \textbf{46.12} & \textbf{22.46} & \textbf{42.92} \\
\bottomrule
\end{tabular}
\caption{\label{tab:sys} Summary level combination on \texttt{CNNDM}.
\textbf{Two} denotes two-system combination (BART and pre-trained \modelname). \textbf{Three} denotes three-system combination (BART, pre-trained \modelname and GSum).
R-1, R-2 and R-L denote ROUGE-1, ROUGE-2 and ROUGE-L. \dag: significantly better than the best single system ($p < 0.01$).}
\end{table}

\subsection{Exp-II: Multiple Systems Stacking}
\label{subsec: stacking}

\begin{table}[t!]
\small
\centering
\begin{tabular}{llll}
\toprule
\textbf{System} & \textbf{R-1} & \textbf{R-2} & \textbf{R-L}\\
\midrule
 BART & 44.26 & 21.12 & 41.16 \\
 Refactor & 44.13 & 20.51 & 40.29 \\
 Min & 31.51 & 10.83 & 28.87\\
 Max & 50.91 & 26.07 & 46.97\\
 Random & 41.66 & 18.77 & 38.27\\
 BERTScore & 43.55 & 20.14 & 39.84 \\
 RankSVM & 43.18 & 19.91 & 39.51 \\
 \midrule
 \singlestep\dag & \textbf{44.96} & \textbf{21.50} & \textbf{41.43} \\
 Pre-trained\dag & 44.88 & 21.13 & 41.16 \\
 Fine-tuned\dag & 44.93 & 21.48 & 41.42 \\
\bottomrule
\end{tabular}
\caption{\label{tab:sent} Sentence level combination on \texttt{CNNDM}.
R-1, R-2 and R-L denote ROUGE-1, ROUGE-2 and ROUGE-L. \dag: significantly better than the best single system ($p < 0.01$).}
\end{table}

\paragraph{Summary-level}
For summary-level combination, we explore two-system combination (BART \& pre-trained \modelname) and three-system combination (BART, GSum \& pre-trained \modelname). 
The results are shown in Tab.~\ref{tab:sys}.

\paragraph{Sentence-level}
For sentence-level combination, 
we use BART and pre-trained \modelname as the base systems. 
The sentences of each system's output are merged together to form the candidate sentence set, and all combinations of three sentences in the candidate set are generated as candidate summaries. 
To prune the candidates, we use tri-gram blocking to filter out candidates of which there exists an identical tri-gram in two sentences. 
The average number of candidates in the test set is 15.8. The results are shown in Tab.~\ref{tab:sent}.

We have the following observations: (1) the pre-trained \modelname can already outperform the base systems, and (2) fine-tuning can further improve the performance. 
Meanwhile, we notice there are two exceptions:
(i) For sentence-level combination, supervised \modelname has similar performance as fine-tuned \modelname.
We hypothesis that this is because here the number of candidates in the fine-tuning data is relatively large, therefore directly training on the fine-tuning data is sufficient enough.
(ii) The pre-trained \modelname cannot outperform GSum model in the three-system combination setting in Tab.~\ref{tab:sys}. 
The reason might be that GSum has much stronger performance than the other two systems, which intuitively makes the expected gain from system combination lower than other settings. 

\subsection{Exp-III: Generalization on 19 Top-performing Systems}
\label{subsec:exp-19}

\begin{table}[t!]
\centering
\small
\begin{tabular}{lcllllll}
\toprule
\textbf{bin}&\textbf{\#sys}&\textbf{Max}&\textbf{Min}&\textbf{Rand}&\textbf{Best}&\textbf{Ours} \\
\midrule
39-40&3&45.28&34.30&39.88&39.98&\textbf{40.45}\\
41-42&8&50.14&32.65&41.44&41.89&\textbf{43.20}\\
42-43&3&47.37&36.79&42.10&42.27&\textbf{43.38}\\
43-44&2&47.60&39.63&43.58&43.97&\textbf{44.07}\\
44-45&3&50.29&38.66&44.58&44.68&\textbf{45.29}\\
\bottomrule
\end{tabular}
\caption{\label{tab:multi} Multiple system combination.
\textbf{bin} denotes the bin range. \textbf{\#sys} denotes the number of systems.
\textbf{Ours} denotes the pre-trained \modelname~model. \textbf{Best} denotes the candidate system with best performance.}
\end{table}

\begin{table*}[htp]
\centering
\begin{tabular}{llllclllclll}
\toprule 
\multirow{2}{*}{\textbf{Method}} & \multicolumn{3}{c}{\textbf{XSum}}  & & \multicolumn{3}{c}{\textbf{PubMed}} & & \multicolumn{3}{c}{\textbf{WikiHow}}\\
\cmidrule{2-4} \cmidrule{6-8} \cmidrule{10 - 12}
& \textbf{R-1} & \textbf{R-2} & \textbf{R-L} & & \textbf{R-1} & \textbf{R-2} & \textbf{R-L} & & \textbf{R-1} & \textbf{R-2} & \textbf{R-L} \\
\midrule
 Base & 47.12 & 24.46 & 39.04 & & 43.42 & 15.32 & 39.21 & & 41.98 & 18.09 & 40.53 \\
 Min & 42.45 & 20.50 & 35.19 & & 39.60 & 13.57 & 35.53 & & 40.55 & 17.40 & 39.18 \\
 Max & 51.51 & 28.04 & 42.70 & & 45.23 & 16.72 & 40.67 & & 43.00  & 18.44 & 41.44 \\
 Random & 46.98 & 24.08 & 38.88 & & 42.39 & 15.12 & 38.08 & & 41.77 & 17.92 & 40.33 \\
 BERTScore & 47.13 & 24.04 & 38.89 & & 43.64 & 15.40 & 39.41 & & 41.77 & 17.93 & 40.29\\
 RankSVM & 46.85 & 24.31 & 39.09 & & 43.63 & 15.34 & 39.46 & & 42.00 & 18.08 & 40.57 \\
\midrule
 Pre-trained & \textbf{47.45} & \textbf{24.55} & \textbf{39.41} & & 43.58 & 15.36 & 39.38 & & 41.97 & 18.03 & 40.52\\
 Fine-tuned & 47.32 & 24.31 & 39.22 & & \textbf{43.72} & \textbf{15.41} & \textbf{39.51} & & \textbf{42.12} & \textbf{18.13} & \textbf{40.66}\\
\bottomrule
\end{tabular}
\caption{\label{tab:single-other} Single system reranking on other datasets.
\textbf{Pre-trained} denotes the pre-trained \modelname~model. \textbf{Fine-tuned} denotes the fine-tuned model. R-1, R-2 and R-L denote ROUGE-1, ROUGE-2 and ROUGE-L separately.}
\end{table*}

To evaluate the \modelname's generalization ability, we explore another setting where the pre-trained \modelname~is directly used to select the outputs of multiple systems without fine-tuning.

To this end, we collect 19 top-performing summarization systems on \texttt{CNNDM} dataset. 
Here, we investigate if our \modelname~can boost the performance of candidate systems with similar performance.
In addition, we also aim to investigate how the range width of different systems' performance affects \modelname's performance. 
Therefore, we group the candidate systems into equal-width bins based on their average ROUGE-1 scores, and evaluate our \modelname~on each bin separately. 

In Tab.~\ref{tab:multi} we report the average ROUGE-1 scores of the oracles, \modelname, and the best candidate system in each bin whose width is 1. \modelname consistently outperforms the best candidate system, showing its generalization ability.  

\begin{figure}
    \centering
    \includegraphics[width=0.65\linewidth]{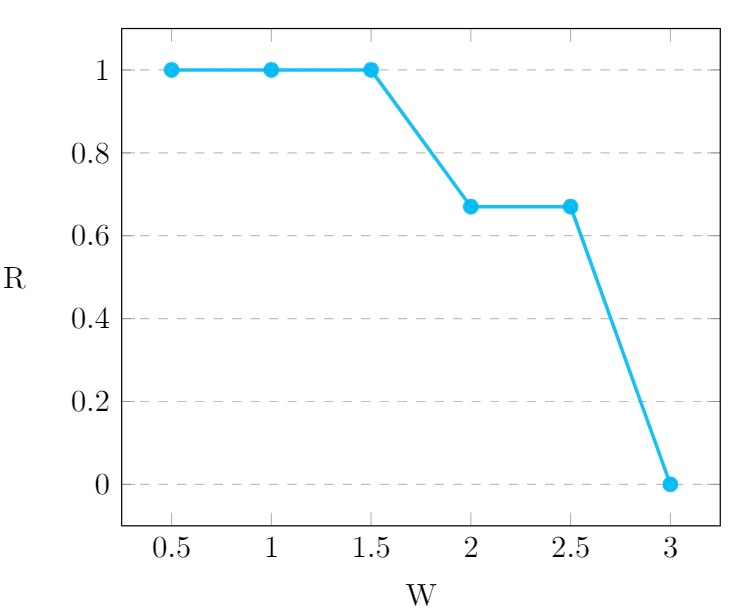}
    \caption{The \modelname's success rates with different bin widths. \textbf{W} denotes the bin widths measured by ROUGE-1. \textbf{R} denotes the success rate of the \modelname outperforming the single best base system.} 
    \label{fig:bin}
\end{figure}

Next, in Fig.~\ref{fig:bin} we plot the change of \modelname's performance with different bin widths. 
We define the success rate of \modelname~with a given bin width to be the number of bins where \modelname~outperforms the single best base system normalized by the total number of bins. 
We observe that \modelname is more likely to improve the performance of base systems when the \textit{system-level} performance of the base systems is similar.
Intuitively, if one base system is significantly better than the other systems, it is more difficult for \modelname to use other systems to complement the best base system.

\subsection{Exp-IV: Effectiveness on More Popular Datasets}

Next, we move on to other text summarization datasets to evaluate our proposed method's strength beyond \texttt{CNNDM} dataset. 
Some of the datasets used here are not as well-studied as \texttt{CNNDM} dataset, so there are less top-performing systems on these datasets.
Therefore, here we focus on the experiments of the single system setting.

\paragraph{Setup}
Regarding the pre-trained \modelname, we use an extractive oracle to select document sentences and use the combinations of these sentences as candidates.
In addition, since on \texttt{Xsum} the abstractive systems outperform extractive systems by a large margin, we use a pre-trained BART model with Diverse Beam Search~\citep{vijayakumar2018diverse} to generate 16 candidates per sample for pre-training.
Regarding system re-ranking, we use BART as the base system to generate the candidate summaries except on \texttt{Xsum} dataset, where we use PEGASUS since it achieves better performance. 
Similar to \S\ref{subsec:cnndm-single}, we use the outputs of beam search as the candidates. 
We select the first 4 outputs as the candidates.

The results in Tab.~\ref{tab:single-other} show that \modelname~is able to bring stable improvement over the base systems. 
The average summary length of these datasets varies from 23.3 (\texttt{XSum}) to 210.3 (\texttt{Pubmed}).
Therefore, the results here demonstrate the \modelname can be applied to datasets with different characteristics.
On \texttt{XSum} dataset, the pre-trained \modelname outperforms the fine-tuned \modelname.
This may result from the additional pre-training data we introduced using BART, which is effective enough to train the \modelname for reranking PEGASUS output.

\subsection{Fine-grained Analysis}

We perform a fine-grained evaluation of \modelname~to understand where improvement mainly comes.
\paragraph{Setup}
We choose the summary-level system combination setting on \texttt{CNNDM} test set in \S\ref{subsec: stacking} as a case study, where the base systems are:  BART and pre-trained \modelname, and then we use a fine-tuned \modelname\footnote{As introduced in \S\ref{sec:scenario}, Refactor could be used as either a base system or a system combinator.} to combine them. 
Specifically, we first

\noindent (i)  define $\delta(C_{\text{BART}}, C_{\text{Pretrain}})$ as the performance (i.e., ROUGE) gap on the candidate summary $C$.

\noindent (ii) then partition test samples into different buckets $S_1, \cdots, S_n$ according to the performance gap $\delta$.

\noindent (iii) calculate \textit{selection accuracy} for each bucket, which represents how accurately the \modelname can identify the best one from two candidate summaries.

\begin{figure}
    \centering
    \includegraphics[width=0.65\linewidth]{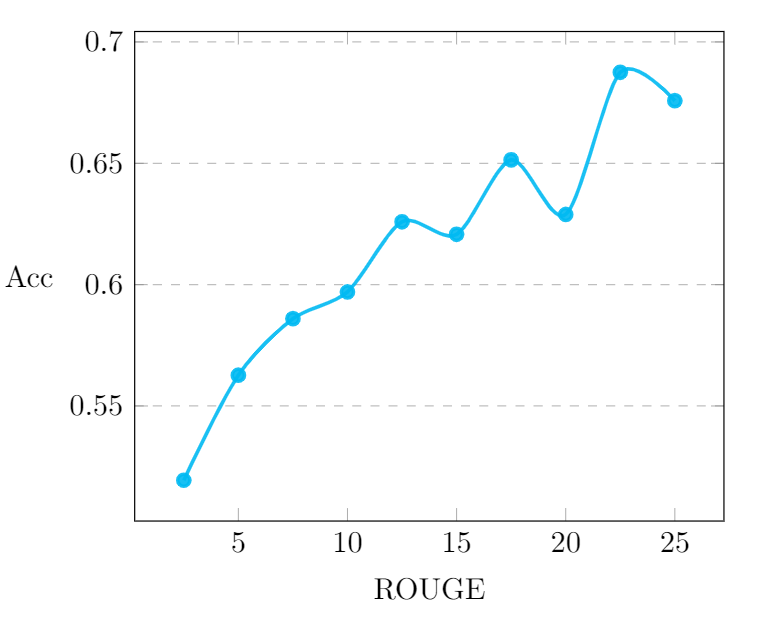}
    \caption{Fine-tuned \modelname's selection accuracy on \texttt{CNNDM} with different difficulties. The X-axis is the difference of ROUGE score of BART and pre-trained \modelname outputs.} 
    \label{fig:acc}
\end{figure}

The results are shown in Fig.~\ref{fig:acc}.
We observe that the selection accuracy is increasing as the gap $\delta$ becoming larger, indicating that \modelname~performs better on the candidate summaries with diverse performance. 
Combining the results we get in \S\ref{subsec:exp-19}, we conclude that \modelname has the largest potential gain when the base systems effectively complement each other -- They have similar \textit{system-level} performance but diverse \textit{summary-level} performance.
For example, each base system may perform significantly better than others on a subset of data with different characteristics but could not outperform others across the whole dataset.

\label{subsec:qual}

\section{Implications and Future Directions}
We present a general framework for utilizing the complementarity of modern text summarization systems by formulating text summarization as a two-stage learning problem.
Our proposed model, \modelname, can be used either as a base system or a meta system, effectively mitigating the learning gaps introduced in the two-stage learning.
Experimental results show that \modelname is able to boost the performance of the base systems, and achieves the state-of-the-art performance on \texttt{CNNDM} and \texttt{XSum} datasets.
We believe this work opens up a new direction for improving the performance of text summarization systems apart from an iterative process of searching for better model architectures -- The gain of performance could be made by fully investigating and utilizing the complementarity of different systems with various architectures, problem formulations, decoding strategies, etc.


\section*{Acknowledgements}

We thank Professor Graham Neubig and anonymous reviewers for valuable feedback and helpful suggestions.
This work was supported in part by a grant under the Northrop Grumman SOTERIA project and the Air Force Research Laboratory under agreement number FA8750-19-2-0200. The U.S. Government
is authorized to reproduce and distribute reprints for Governmental
purposes notwithstanding any copyright notation thereon. The views and
conclusions contained herein are those of the authors and should not be
interpreted as necessarily representing the official policies or
endorsements, either expressed or implied, of the Air Force Research
Laboratory or the U.S. Government.

\bibliography{anthology,custom}
\bibliographystyle{acl_natbib}

\appendix

\section{Features for RankSVM}
\label{sec:ranksvm}

We use 18 features as defined below for RankSVM:
\begin{enumerate}
    \item document length.
    \item candidate summary length.
    \item rouge-1, rouge-2, rouge-L between source documents and candidates summaries.
    \item copy length: the length of summary's fragments appeared in the source document.
    \item fragment coverage, fragment density, compression ratio as defined in \citet{grusky-etal-2018-newsroom}.
    \item novelty: the ratio of novel $k$-grams ($k \in \{1, 2, 3, 4\}$) in the candidate summaries.
    \item repetition:  the ratio of repeated $k$-grams ($k \in \{1, 2, 3, 4\}$) in the candidate summaries.
    \item sentence fusion ratio: the ratio of sentences in the candidate summaries that combine the content of two source document sentences.
\end{enumerate}

\end{document}